\documentclass[10pt,twocolumn,letterpaper]{article}

\usepackage{wacv}              

\usepackage{graphicx}
\usepackage{amsmath}
\usepackage{amssymb}
\usepackage{booktabs}

\usepackage[pagebackref,breaklinks,colorlinks]{hyperref}

\usepackage[capitalize]{cleveref}
\crefname{section}{Sec.}{Secs.}
\Crefname{section}{Section}{Sections}
\Crefname{table}{Table}{Tables}
\crefname{table}{Tab.}{Tabs.}

\def\wacvPaperID{2177} 
\def\confName{WACV}
\def\confYear{2025}

\title{IP-FaceDiff: Identity-Preserving Facial Video Editing with Diffusion}

\author{
Tharun Anand$^{1}$, Aryan Garg$^{2}$, Kaushik Mitra$^{1}$\\
$^{1}$Indian Institute Of Technology Madras, $^{2}$University of Wisconsin Madison\\
{\tt\small \{ed20b068, kmitra\}@smail.iitm.ac.in, agarg54@wisc.edu}
}

\begin{document}

\twocolumn[{
\maketitle
\begin{center}
    \captionsetup{type=figure}
    \includegraphics[width=0.95\textwidth,height=0.35\textheight]{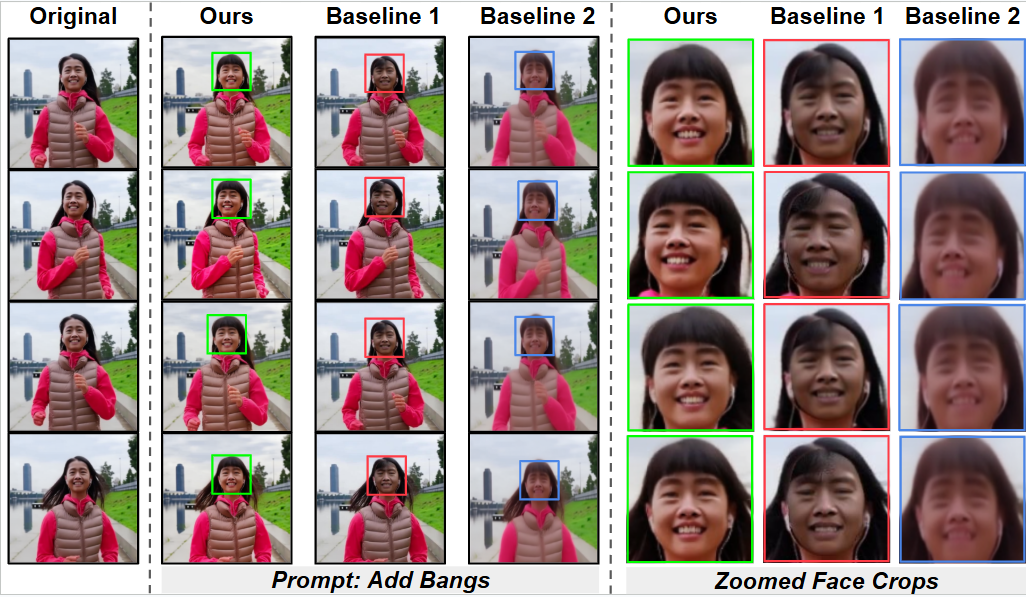}
    \caption{\textbf{Qualitative Facial Attribute Editing.} Our method enables precise, localized facial edits whilst maintaining low computational latency. Through explicit identity preservation optimization, we ensure that global facial identity features are retained. Furthermore, our approach ensures temporal consistency and robust generalization across a wide range of facial video editing tasks.}
\label{fig:start}
\end{center}
}]

\begin{abstract}
\vspace{-0.5 cm}
Facial video editing has become increasingly important for content creators, enabling the manipulation of facial expressions and attributes. However, existing models encounter challenges such as poor editing quality, high computational costs and difficulties in preserving facial identity across diverse edits. Additionally, these models are often constrained to editing predefined facial attributes, limiting their flexibility to diverse editing prompts. 
To address these challenges, we propose a novel facial video editing framework that leverages the rich latent space of pre-trained text-to-image (T2I) diffusion models and fine-tune them specifically for facial video editing tasks. Our approach introduces a targeted fine-tuning scheme that enables high quality, localized, text-driven edits while ensuring identity preservation across video frames. Additionally, by using pre-trained T2I models during inference, our approach significantly reduces editing time by  80\%, while maintaining temporal consistency throughout the video sequence. We evaluate the effectiveness of our approach through extensive testing across a wide range of challenging scenarios, including varying head poses, complex action sequences, and diverse facial expressions. Our method consistently outperforms existing techniques, demonstrating superior performance across a broad set of metrics and benchmarks.
\end{abstract}

\vspace{-5 mm}
\section{Introduction}
\label{sec:intro}

Digital content generation and editing have experienced remarkable strides since the advent of diffusion models~\cite{ddim,ddpm,10.5555/3600270.3600688,ho2022classifierfree}. 
These generative models have been progressively trained to enhance input image clarity but often lack precise control over the output. 
However, recent advancements in conditional diffusion models~\cite{Meng2021SDEditGI,NEURIPS2021_49ad23d1} have addressed this limitation, opening up exciting avenues for multi-modal or text-controlled generation~\cite{gafni2022makeascene,saharia2022photorealistic,ramesh2022hierarchical,rombach2022highresolution,Chefer2023AttendandExciteAS}. 
These innovative approaches have demonstrated effectiveness and improved quality across various tasks, including image generation~\cite{NEURIPS2021_49ad23d1,Graikos_2023_BMVC,Hwang_2023_CVPR}, diverse image editing scenarios~\cite{Brooks2022InstructPix2PixLT,hertz2022prompttoprompt,gal2022image}, and even video editing applications~\cite{qi2023fatezero,molad2023dreamix,Chai2023StableVideoTC}.

Facial video editing differs from conventional video editing, where changes are often applied to broader scenes. Our approach focuses on the unique challenges of facial editing, such as executing precise localized modifications, maintaining temporal consistency across frames, and preserving the original video's quality and subject’s identity throughout the process. These challenges demonstrate why conventional video editing models are not suited for the precision required in facial video editing.
Previous works in GAN-based~\cite{tzaban2022stitch, Abdal2019Image2StyleGANHT,yao2021latenttransformerdisentangledface} and Diffusion based~\cite{Kim2022DiffusionVA} facial video editing faced similar hurdles. They often produced visible artifacts and also handled each frame independently, which significantly increased computation cost and editing time. Moreover, these techniques require end-to-end training tailored to specific facial videos and editing prompts, limiting their ability to generalize to diverse prompts and unseen head poses.

\par
To tackle these challenges, we employ pre-trained Text-to-Image (T2I) diffusion models (Stable Diffusion 2.1 ~\cite{podell2023sdxl}), known for its generalization capabilities and high-quality image generation capability due to large-scale pre-training. Also using pretrained T2I models for Text2video tasks has proven to achieve significant reductions in editing time and compute latency~\cite{geyer2023tokenflow,wu2023tuneavideo} . 

While off-the-shelf (T2I) models provide strong generalization to unseen faces, they often struggle with precision when adhering to localized prompts and preserving identity~\cite{geyer2023tokenflow,wu2023tuneavideo} Figure \ref{fig:start}.
To address this, we use directional-CLIP loss~\cite{Gal2021StyleGANNADA} to fine-tune one branch of our parallel T2I framework to enable it to make highly localized edits to the facial video.
Additionally, we integrate an identity preservation loss~\cite{Deng_2022}, into one of our fine-tuned T2I branches. 
This explicitly guides the editing process during \textit{inference}, focusing on maintaining the person's identity.
\par
The combination of these two independently fine-tuned diffusers \& careful loss setups enables our method to perform identity preserving localized edits to facial videos(\cref{fig:multiple_edits}).

In summary, we make the following contributions:
\begin{itemize}
    \item We present a novel pipeline for facial video editing that leverages pre-trained text-to-image (T2I) diffusion models. This approach enables high quality prompt-consistent edits, generalizes effectively to challenging in-the-wild facial videos, and significantly reduces both computational cost and latency.
    \vspace{-1mm}
    \item We propose an inference-time guidance strategy that preserves the original identity throughout the editing process, ensuring global identity consistency across frames.
    \vspace{-1mm}
    \item We develop a fine-tuning strategy that allows for highly localized edits in facial videos and demonstrate that this framework significantly surpasses existing benchmarks for facial video editing in various quantitative and qualitative aspects.
\end{itemize}

\section{Related Work}
\label{sec:related}
\subsection{Text to Image Diffusion models} 
 Text-guided diffusion models encode the rich latent features of an image coupled with the text guidance through cross-attention modules~\cite{Meng2021SDEditGI,saharia2022photorealistic,ramesh2022hierarchical,gafni2022makeascene}. ControlNet~\cite{controlNet}, LORA~\cite{hu2022lora} and PNP-diffusion~\cite{tumanyan2022plugandplay} delve into the controllability of visual generation by incorporating additional encoding layers, facilitating controlled generation under various conditions such as pose, mask, edge and depth. Notably ~\cite{tumanyan2022plugandplay} 
 controls image generation by replacing the self-attention features of selective layers of the T2I network with another network that effectively preserves global features of the image and \cite{controlNet} provides guidance through zero convolutions. We take inspiration from these methods to facilitate additional identity guidance to our editing method. 

\subsection{Text guided video editing} Unlike advancements in image manipulation, progress in video manipulation methods has been slower, with a key challenge being the need to streamline editing processes for practical use, while ensuring temporal consistency across frames. Previous efforts ~\cite{qi2023fatezero,molad2023dreamix,Khachatryan2023Text2VideoZeroTD} have tried to address these challenges but often at the expense of computational cost and impractical editing times.
\par
To address this, recent advances~\cite{geyer2023tokenflow,wu2023tuneavideo,Ceylan2023Pix2VideoVE} demonstrated using pre-trained T2I diffusion models results in faster video editing. In particular Geyer~\etal~\cite{geyer2023tokenflow} proposed using a T2I model with keyframe editing instead of editing every frame. Changes in these keyframes are propagated to other frames via interframe correspondences, significantly reducing editing time. Yao~\etal~\cite{wu2023tuneavideo} leveraged a pretrained (T2I) diffusion model and fine-tunes it with a small number of iterations using latent code initialization and temporal attention, significantly reducing video editing time.

\subsection{Facial Video Editing}
Facial video editing presents a unique challenge compared to traditional methods. Here, preserving the original person's identity and natural head movements is crucial, while maintaining temporal coherence throughout the edited video . Early methods~\cite{tzaban2022stitch,Abdal2019Image2StyleGANHT,9157575,Bau_2019,Gal2021StyleGANNADA,Gu2019ImagePU} leveraged StyleGAN's well-structured latent space for facial video editing. By manipulating specific codes, they could alter facial features like expressions, gender, and age. However, they were limited to broader edits and couldn't address more localized details like beards, glasses, or accessories.
\par
 To better preserve temporal consistency while allowing for more localized edits to the facial video, Kim~\etal~\cite{Kim2022DiffusionVA} proposed to use DDIM~\cite{ddim} to edit facial videos. 
This method first encodes the frames of a facial video into time-invariant (identity) and time-variant (motion) features using separate pre-trained encoders, then performs CLIP-guided editing with the text prompt to the time-invariant features. 
Then along with the noisy latents of the facial frames (obtained from the deterministic forward process of DDIM) and these edited features, they guide the reverse process of the conditional DDIM. 
But this involves a lot of individual preprocessing steps and processes every frame of the video during inference (\cref{tab:inference}). 
In contrast, our method tackles these challenges through a video editing framework that leverages off-the-shelf T2I models. To the best of our knowledge we are the first to explore using pretrained T2I models for facial video editing tasks. 

\section{Preliminary}
\subsection{Denoising Diffusion Probabilistic Models:} 
DDPMs~\cite{ddpm, ddim} are generative models based on iterative Markovian steps: forward noising and reverse denoising. Given an input image $y_0$, the forward process progressively adds noise using a Markov chain $q$ over $T$ steps:
\begin{equation}
q(y_{1:T} \mid y_0) = \prod_{t=1}^T q(y_t \mid y_{t-1}) = \mathcal{N}(y_t \mid \sqrt{\alpha_t} y_{t-1}, (1-\alpha_t) I)
\end{equation}
where $\alpha_t$ controls the added noise level at each step. The reverse process uses a U-Net $\epsilon(y_t, t)$ to predict and remove the noise at each step, progressively denoising $y_t$ back to $y_0$. This process can be guided by additional input signals, such as text descriptions or reference images, allowing the model to generate outputs that are more specific and aligned with the given conditions. By incorporating these inputs, the generative process becomes highly flexible and can be tailored to produce desired outputs that match particular requirements or constraints.

We leverage a pre-trained text-conditioned Latent Diffusion Model (LDM), like Stable Diffusion~\cite{podell2023sdxl} for our task, which applies the diffusion process to a pre-trained image autoencoder's latent space. The U-Net architecture uses a residual block, a self-attention block to capture long-range dependencies and a cross-attention block for image features's interaction with text embeddings from the conditioning prompt.

\subsection{Joint Keyframe Editing:} 
\label{subsec:joint_keyframe_editing}
Traditional video editing techniques necessitate editing each frame separately and then ensuring temporal coherence, leading to significant computational overhead. 
To mitigate this, Geyer et al.~\cite{geyer2023tokenflow} exploit the temporal redundancies within the feature space of T2I diffusion models in natural videos. 
This approach enables the attainment of temporal coherence in editing by altering only a selected subset of keyframes and propagating these modifications throughout the video.

 The keyframe $k_i$ generates a feature representation $f_i$, followed by the calculation of an attention weight $a_{ij}$ for $k_i$, indicating the relevance of information with another keyframe $k_j$ as follows:
\begin{equation}
    a_{ij} = \text{softmax}(f_i^T W_a f_j)
    \label{eq:a_ij}
\end{equation}
where $W_a$ is a learned weight matrix. 
After computing attention weights, each keyframe's feature is updated by aggregating information from all other keyframes:


\begin{equation}
    f_i^{\text{updated}} = f_i + \sum_j a_{ij} W_c f_j
    \label{eq:fi_updated}
\end{equation}
where $W_c$ is another learned weight matrix. 
Then For each frame $i$ they proceed to find the nearest neighbors in the original video's keyframe feature space. 
Let $\gamma_{i^+}$ and $\gamma_{i^-}$ denote the indices of the closest future and past keyframes (respectively) to frame $i$.
Edited keyframes features ($T_{\text{base}}$) are propagated to $f_i$ based on its distance to neighboring keyframes:
\begin{equation}
    f_i^{\text{prop}} = w_i^+ T_{\text{base}}(\gamma_i^+) + w_i^- T_{\text{base}}(\gamma_i^-)
    \label{eq:f_propagation}
\end{equation}
where $T_{\text{base}}(\gamma_{ij})$ refers to the edited feature vector of the keyframe with index $\gamma_{ij}$ (either future or past), and $w_{i^+}$ and $w_{i^-}$ are weights determined by the distance between frame $i$ and its corresponding neighbors.
This technique ensures consistent video editing by integrating style information from all keyframes simultaneously, fostering global-editing coherence.

\begin{figure*}[ht]
    \centering
    \includegraphics[width=0.99\linewidth]{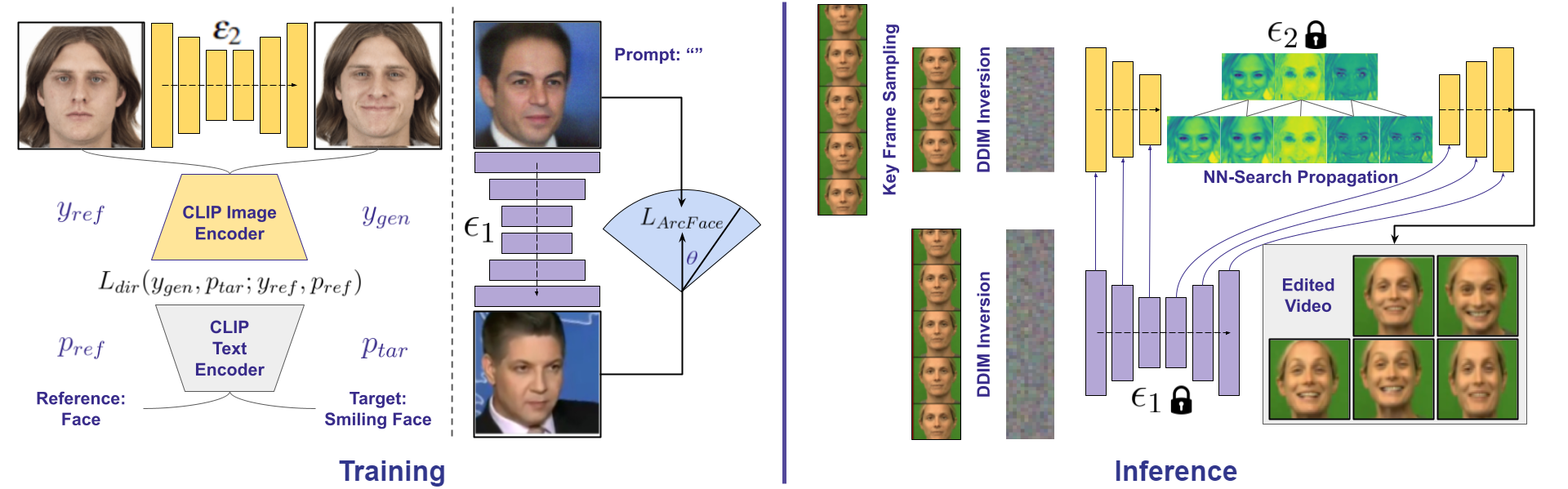}
    \caption{\textbf{Model Architecture.} Left: Pre-trained T2I models $\epsilon_1$ and $\epsilon_2$ are fine-tuned independently with ArcFace loss and directional CLIP loss for identity-preservation and prompt-adhering localization, respectively. Right: Video frames are inverted with DDIM and then processed through $\epsilon_1$ to extract self-attention features at each timestep. Text-guided editing is applied to keyframes using $\epsilon_2$, guided by identity features from $\epsilon_1$. Edits are propagated to remaining frames using a nearest-neighbor search within the latent space.}
    \label{fig:model}
\end{figure*}

\section{Method}
\label{sec:method}
This section details our facial video editing approach, consisting of two main stages: preprocessing and editing. 
In the preprocessing phase, identity features are extracted from video frames using the pre-trained T2I model $\epsilon_1$. For each frame \(x_{i}\) in the input video (with \(i\) ranging from 1 to \(N\) frames), a DDIM inversion process is applied with $\epsilon_1$ to extract self-attention features \(x_{l}^{i}\) at every network layer \(l\) during denoising, as depicted in Figure \textcolor{red}{2}. In the editing step, we adopt joint keyframe editing(~\cref{subsec:joint_keyframe_editing}), focusing text-guided edits on sampled keyframes \(x_{0}^{k}\) (where \(k\) ranges from 1 to \(N\)) and then propagating these changes to the remaining frames using a nearest neighbor search~\cite{geyer2023tokenflow} with network $\epsilon_2$.
\par
To ensure identity preservation, we introduce a novel pipeline where during editing we substitute the self-attention features of $\epsilon_2$ with those precomputed by $\epsilon_1$. 
Additionally, we've devised a framework to fine-tune $\epsilon_2$ for more precise localized facial edits.

\subsection{Inference time Identity Preservation}
Naively applying Text2Video models~\cite{molad2023dreamix,geyer2023tokenflow,Khachatryan2023Text2VideoZeroTD} for facial video editing often results in a loss of the original person's identity. 
These models are trained for generic video editing tasks such as changes in global style, scene of an image and thus when applied for facial videos struggle to  maintain critical facial features.

To tackle this challenge, we introduce a method that incorporates additional structure and identity guidance during the editing process.
\par
We fine-tuned an off-the-shelf T2I model $\epsilon_1$ using a large facial image dataset (CelebA HQ~\cite{xia2021towards}), employing a combined loss function that minimizes both pixel-wise reconstruction errors and identity differences (using ArcFace~\cite{Deng_2022}) between generated and ground truth images. 
This approach aids in face reconstruction during editing while preserving identity.
Mathematically we reduce the following loss:
\begin{equation}
L_{id} = \| \epsilon_1(x_{0}) - y_{0} \|_1 +  D_{\cos}\left(f_{arc}(\epsilon_1(x_{0})), f_{arc}(y_{0})\right)
\label{eq:Losses}
\end{equation}
where \(y_{0}\) is the ground truth face image and \(x_{0}\) is the noisy latent obtained by DDIM inversion of \(y_{0}\).
The second term minimizes a cosine-similarity identity Loss using the pre-trained  ArcFace~\cite{Deng_2022} network (\(f_{arc}\)) where \(D_{\cos}\) represents the cosine distance between the two features.

\par
 During inference, we guide the editing process of $\epsilon_2$ by replacing its self-attention features with those from corresponding layers of the identity-preserving $\epsilon_1$ network.
This approach effectively incorporates identity information into the editing process, resulting in robust preservation of facial identity and head pose in the edited videos, as in Figure \textcolor{red}{2}.

\subsection{Directional Editing}
T2I diffusion, pre-trained on generic generative tasks like style transfer and scene editing, are not well-suited for localized editing and often end up changing global semantics while increasing artifacts.
TokenFlow~\cite{geyer2023tokenflow}, a non-faces geared (generic) diffusion-based video editor, suffers from this issue, see~\cref{fig:start}.
To address this limitation, we fine-tune a pre-trained T2I model (SD 2.1~\cite{podell2023sdxl}) using a directional CLIP loss~\cite{Gal2021StyleGANNADA} on the large-scale face image dataset: CelebA-HQ~\cite{xia2021towards}. 
We minimize the loss defined by:
\begin{align}
\mathcal{L}_{dir}(\mathbf{y_{gen}}, \mathbf{p_{tar}}) &= \lambda_1 \| f_I(\mathbf{y_{gen}}) - f_I(\mathbf{y_0}) \|^2 \nonumber \\
&\quad + \lambda_2 \left[1 - \left( \left(f_I(\mathbf{y_{gen}}) - f_I(\mathbf{y_{ref}})\right) \cdot \right. \right. \nonumber \\
&\quad \quad \left. \left. \left(f_T(\mathbf{p_{tar}}) - f_T(\mathbf{p_{ref}})\right)\right) \right]
\end{align}

\noindent where $\mathcal{L}_{dir}(\mathbf{y_{gen}}, \mathbf{p_{tar}})$ is the directional CLIP loss for image $\mathbf{y_{gen}}$ and target prompt $\mathbf{p_{tar}}$. 
$f_{I/T}(\cdot)$ denotes the CLIP Image/Text encoder mapping images and text into the same latent space. $\mathbf{y_{ref}}$ is original image. 
$\mathbf{p_{ref}}$ is the reference text embedding.
$\lambda_1$ and $\lambda_2$ are scalar hyper-parameters.

\begin{figure*}[ht]
    \centering
    \includegraphics[width=0.99\linewidth]{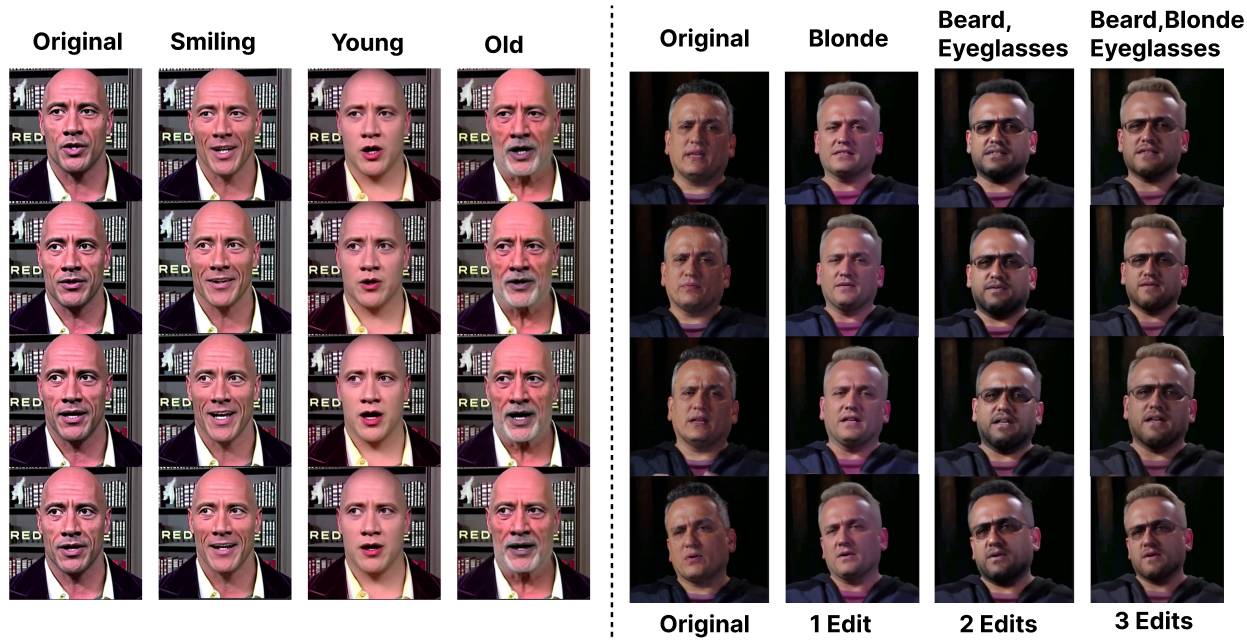}
    \caption{\textbf{Strong Prompt-Adhering Multiple Editing.} \textit{Left}: Beyond local edits, our method manipulates facial expressions and age. \textit{Right}: Our facial video editing method handles simultaneous edits to local facial features.}
    \label{fig:multiple_edits}
\end{figure*}
The directional CLIP loss~\cite{Gal2021StyleGANNADA} effectively guides the editing process by minimizing the directional change between the original facial image representation and the text prompt's influence in model's latent space. 
This loss allows for more localized edits compared to traditional clip losses.

We call this fine-tuned T2I model as $\epsilon_2$ Figure \textcolor{red}{2}. During fine-tuning, we focused on 2 edit categories: (1) \textit{facial features} like age, gender and expressions and (2) \textit{facial attributes} like hair color, spectacles, mustaches, or beards.

\section{Experiments}
\subsection{Implementation Details}
We use Stable Diffusion 2.1~\cite{podell2023sdxl} as our pre-trained Text-to-Image model ($\epsilon_1$) for fine-tuning both our branches. 
We fine-tune $\epsilon_1$ using the CelebAHQ dataset~\cite{CelebAMask-HQ} comprising of 512$\times$512 10,000 images. 
$\epsilon_2$ was finetuned on the same dataset using captions generated by BLIP-2~\cite{li2023blip} as reference prompts. These were paired with extensive set of editing prompts that covered both facial attributes (e.g., beard, glasses) and facial features (e.g., gender, age). We ensured prompt diversity by incorporating a wide range of facial editing categories present in the large-scale facial attribute dataset CelebA~\cite{liu2015faceattributes}.
 The hyperparameters for the diffusion CLIP loss, $\lambda_1$ is 0.3 and $\lambda_2$ is 0.7.
For both $\epsilon_1$ and $\epsilon_2$, we trained at a resolution of 512x512, batch size of 16 with a learning rate of $10^{-5}$ for 30,000 iterations
using 2 NVIDIA A100's. Inference can be performed on a single one.

\begin{figure*}[ht]
 \centering
 \includegraphics[width=0.99\textwidth,height=0.3\textheight]{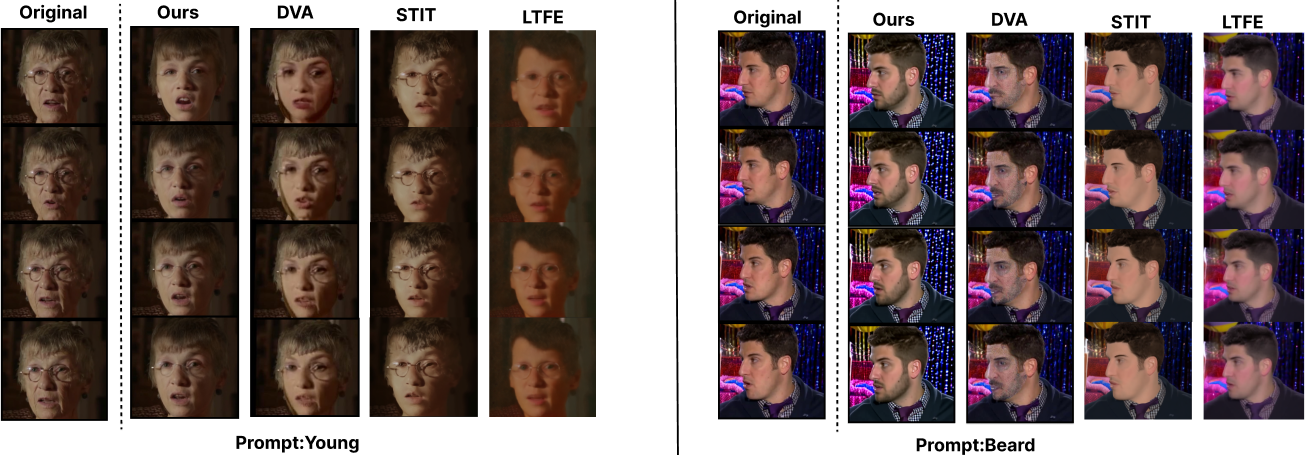}
 \caption{\textbf{Editing Faces in the Wild.} We successfully overcome a previous hurdle of out-of-domain adaptation for facial video editing methods.}
\label{fig:fig_4}
\end{figure*}

\subsection{Baselines and Dataset}
We conducted both quantitative and qualitative experiments, comparing our method against established Facial Video Editing benchmarks:  
\begin{enumerate}  
    \item \textit{DVA}: Kim~\etal's~\cite{Kim2022DiffusionVA} diffusion-based method for facial video editing that enables editing a set of predefined facial features (e.g., expressions, age) and local facial attributes (e.g., beard, accessories).  
    \item \textit{StitchGAN (STIT)}~\cite{tzaban2022stitch}: This method exclusively edits predefined global facial features (e.g., age, gender, expressions) but struggles with localization.  
    \item \textit{Latent Transformer for Facial Video Editing (LTFE)}~\cite{yao2021latenttransformerdisentangledface}: A StyleGAN-based~\cite{Karras2018ASG} facial video editing method limited to a narrow set of attributes.  
\end{enumerate}  

Other recent advancements in facial video editing leveraging StyleGAN inversion have been reported, but experimental comparisons with these models were not feasible due to the lack of publicly available implementations or pretrained models \cite{10012505,Xu2023RIGIDRG,Zhang2024FEDNeRFAH}. Consequently, we focused on widely recognized benchmarks that remain competitive.

\vspace{1\baselineskip}
\textbf{Evaluation Dataset}
For our experiments, we selected a subset of 50 original videos from the CelebV-HQ dataset~\cite{clip2022celebvhq}. From these, we created an evaluation dataset consisting of 25 edited videos of each of the baselines and our method. Each video had a duration of 4 seconds (10 fps) and a frame resolution of 512x512.

\subsection{Identity Preservation Metric}
To quantitatively assess identity preservation in edited videos, we use three face recognition models: CosFace~\cite{cosface}, VGGFace~\cite{Parkhi_2015}, and FaceNet~\cite{Schroff_2015}. These models generate high-dimensional embeddings that capture facial features, enabling identity evaluation.

For each video, we compute embeddings by averaging frame-wise embeddings for the original and edited frames. Identity preservation is assessed using retrieval metrics: 1) Recall at Rank 1 (R@1) \cite{voorhees1999trec}, 2) Mean Reciprocal Rank (MRR) \cite{manning2008ir}, and 3) Cosine Similarity. Using the CelebV-HQ database \cite{clip2022celebvhq} (35,000 videos), we rank the embedding of the original video among the database embeddings based on its similarity to the edited video embedding. 

\noindent\textbf{Recall at Rank 1 (R@1):} Measures the proportion of queries where the original embedding is ranked first based on average Euclidean distance of the frames:
\[
\text{R@1} = \frac{\text{Number of correct top-1 queries}}{\text{Total queries}}
\]

\noindent\textbf{Mean Reciprocal Rank (MRR):} Quantifies retrieval performance by averaging reciprocal ranks of ground-truth embeddings, prioritizing higher ranks:
\[
\text{MRR} = \frac{1}{N} \sum_{i=1}^{N} \frac{1}{\text{rank}_i}
\]
where $\text{rank}_i$ is the position of the ground-truth embedding for the $i$-th query.

\noindent\textbf{Cosine Similarity:} Evaluates similarity between embeddings of original ($x_{\text{0}}$) and edited frames ($x$) to assess identity preservation:
\[
d_{\text{cosine}}(x_{\text{0}}, x) = 1 - \frac{x_{\text{0}} \cdot x}{\|x_{\text{0}}\| \|x\|}
\]
Lower values of $d_{\text{cosine}}$ indicate better identity preservation.

\begin{table}[ht]
 \centering
 \resizebox{\columnwidth}{!}{%
 \begin{tabular}{l c c c c c c c c c} 
 \toprule
 \textbf{Model} & \multicolumn{3}{c}{\textbf{VGGFace~\cite{Deng_2022}}} & \multicolumn{3}{c}{\textbf{CosFace~\cite{cosface}}} & \multicolumn{3}{c}{\textbf{FaceNet~\cite{cosface}}} \\ 
 \cmidrule(lr){2-4} \cmidrule(lr){5-7} \cmidrule(lr){8-10}
 & \textbf{Cosine $\downarrow$} & \textbf{R@1 $\uparrow$} & \textbf{MRR $\uparrow$} & \textbf{Cosine $\downarrow$} & \textbf{R@1 $\uparrow$} & \textbf{MRR $\uparrow$} & \textbf{Cosine $\downarrow$} & \textbf{R@1 $\uparrow$} & \textbf{MRR $\uparrow$} \\ 
 \midrule
 DVA~\cite{Kim2022DiffusionVA} & 0.361 & 0.76 & 0.794 & 0.318 & 0.76 & 0.794 & 0.256 & 0.76 & 0.794 \\ 
 STIT~\cite{tzaban2022stitch} & 0.447 & 0.70 & 0.748 & 0.471 & 0.70 & 0.748 & 0.273 & 0.72 & 0.773 \\ 
 LTFE~\cite{yao2021latenttransformerdisentangledface} & 0.506 & 0.64 & 0.662 & 0.531 & 0.64 & 0.673 & 0.488 & 0.66 & 0.683 \\ 
 \midrule
 {Ours (w/o ID guidance)} & {0.762} & {0.32} & {0.352} & {0.722} & {0.34} & {0.396} & {0.706} & {0.34} & {0.396} \\ 
 \textbf{Ours (with ID guidance)} & \textbf{0.221} & \textbf{0.96} & \textbf{0.97} & \textbf{0.206} & \textbf{0.96} & \textbf{0.97} & \textbf{0.185} & \textbf{0.96} & \textbf{0.97} \\ 
 \bottomrule
 \end{tabular}%
 }
 \caption{\textbf{Quantitative Comparison of Identity Preservation.} We evaluate identity preservation using Cosine Distance, R@1, and MRR metrics averaged across videos in our evaluation dataset, using embeddings from VGGFace, CosFace, and FaceNet.
}
 \label{tab:identity}
\end{table}

\vspace{-1 mm}
\subsection{Frames vs. Inference Time.}
Existing facial video editing benchmarks require end-to-end training for each facial video prior to inference, resulting in high computational costs and significant time consumption. In contrast, our method leverages our pre-trained T2I models to directly perform edits, eliminating the need for additional per-video training. Also by focusing on editing only keyframes, our approach reduces inference time by approximately 80\% compared to state-of-the-art methods, offering a robust \& practical solution for facial editing.

For instance, editing a 30-second, 10 fps facial video (300 frames) takes just $\sim$6 minutes on a single NVIDIA A100 GPU, demonstrating significant gains over the nearest baselines: STIT~\cite{tzaban2022stitch} (32 minutes) and DVA~\cite{Kim2022DiffusionVA} (28 minutes). 
\cref{tab:inference} shows a quantitative comparison.
\begin{table}[ht]
\centering
 \resizebox{\columnwidth}{!}{%
 \begin{tabular}{lcccc} 
    \toprule
    \textbf{Video Duration} & \textbf{Ours} & \textbf{Kim}~\cite{Kim2022DiffusionVA} & \textbf{Tzaban}~\cite{tzaban2022stitch} & \textbf{Yao}~\cite{yao2021latenttransformerdisentangledface} \\ 
     \midrule
    30 sec & \textbf{6 mins} & 28 mins & 32 mins & 35 mins  \\ 
    60 sec & \textbf{15 mins} & 45 mins & 60 mins & 68 mins \\
    90 sec & \textbf{32 mins} & 90 mins & 110 mins & 132 mins \\
    120 sec & \textbf{98 mins} & 180 mins & 194 mins & 206 mins\\ 
    \bottomrule
    \end{tabular}%
    }
    \vspace{0.9mm}
    \caption{\textbf{Inference Speed Comparison} on \textbf{10 fps} videos, averaged across video clips from the evaluation dataset and reported in minutes using a single NVIDIA A100
    }
    \label{tab:inference}
\end{table}
\vspace{-3.5mm}

\subsection{Temporal Consistency Evaluation}
Our method incorporates Geyer \etal's~\cite{geyer2023tokenflow} algorithm, which enforces temporal consistency by aligning and propagating latent tokens across frames within the latent space of pre-trained T2I diffusion models. 
\begin{figure*}[ht]
    \centering
    \includegraphics[width=0.99\linewidth]{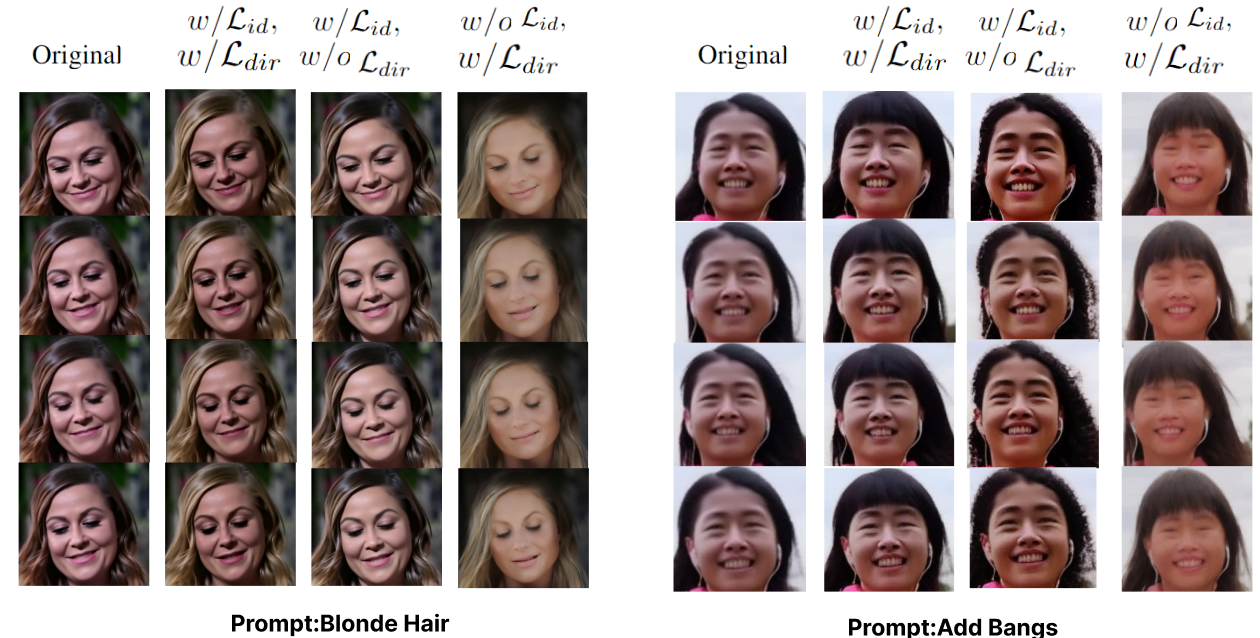}
    \caption{Ablation: fine-tuning $\epsilon_1$ and $\epsilon_2$ with Arc-Face and directional-Clip Loss for identity preservation and performing localized edits in facial videos.}
    \label{fig:abalation}
\end{figure*}

In contrast to existing benchmark approaches that rely on per-frame encoders and process each frame independently, our method inherently enforces frame-to-frame coherence at the latent representation level. This helps in reducing perceptual flickering and improving temporal stability \cref{tab:mos}. 

To quantitatively assess temporal consistency, we utilize the pre-trained RAFT model~\cite{raft_RAFT} to compute optical flow \( o_t \) between consecutive frames \( I_t \) and \( I_{t+1} \). Using the computed optical flow, we forward-warp frame \( I_t \) to frame \( I_{t+1} \) with the bilinear inverse warping operator~\cite{invWarpOperator_NIPS}, denoted as \( \mathcal{W} \). This process ensures precise alignment between consecutive frames, enabling an accurate evaluation of temporal coherence.

The temporal loss, calculated as the average \( \ell_1 \) distance between the warped frame \( \mathcal{W}(I_t, o_t) \) and the actual frame \( I_{t+1} \) across all video frames, is defined as:

\begin{equation}
    \mathcal{L}_{\text{temp}} = \frac{1}{T} \sum_{t=1}^{T} 
    \| \mathcal{W}(I_t, o_t) - I_{t+1} \|_1,
    \label{eq:temp-loss}
\end{equation}

where \( T \) is the total number of frames in the video. A lower temporal loss indicates superior temporal consistency, with smoother transitions and fewer artifacts. \Cref{tab:mos} presents the temporal loss values, comparing our approach against state-of-the-art methods, demonstrating the effectiveness of our method in maintaining temporal coherence across frames.

\begin{figure*}[ht]
    \centering
    \includegraphics[width=0.99\linewidth]{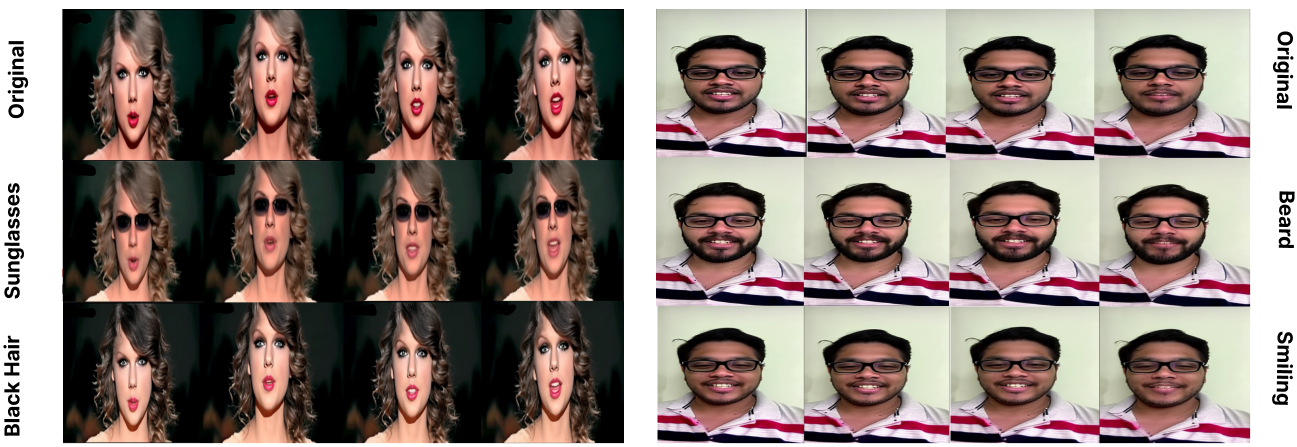}
    \caption{\textbf{More Identity Preserving Localized Editing in the Wild.} Left: Hair color change and accessory addition are performed from a randomly scraped online music video. Right: Facial hair and facial expressions of a consenting volunteer are edited using our method.}
    \label{fig:last_img}
\end{figure*}

\begin{table}[ht]
\centering
\renewcommand{\arraystretch}{1.2} 
\setlength{\tabcolsep}{5pt} 
\begin{tabular}{lccc} 
\toprule
\textbf{Method} & \textbf{Temporal Consistency $\downarrow$} & \textbf{MOS Score $\uparrow$} \\ 
\midrule
DVA~\cite{Kim2022DiffusionVA} & 0.256 & 4.7 \\
STIT~\cite{tzaban2022stitch} & 0.273 & 5.6 \\
LTFE~\cite{yao2021latenttransformerdisentangledface} & 0.463 & 3.4\\
\midrule
\textbf{Ours}  & \textbf{0.238} & \textbf{7.7} \\ 

\bottomrule
\end{tabular}
\vspace{2mm} 
\caption{\textbf{Quantitative Fidelity Comparison.} The table compares identity preservation, RAFT-based temporal consistency, and Mean Opinion Score (MOS) for different methods. Evaluations are conducted on 20 ten-second videos at $512\times512$ resolution, highlighting the superior temporal consistency and perceived quality of our approach compared to existing methods.}
\label{tab:mos}
\end{table}

\subsection{Ablation Study}
We conducted an ablation study to assess the contributions of key components in our model: the effect of identity guidance from the fine-tuned model $\epsilon_1$, optimized with \( \mathcal{L}_{\text{id}} \), and the impact of fine-tuning $\epsilon_2$ with \( \mathcal{L}_{\text{dir}} \) to achieve temporally consistent and localized facial edits.

First, we retained the identity guidance from $\epsilon_1$ and edited facial videos with $\epsilon_2$ before it was fine-tuned with the directional CLIP loss \( \mathcal{L}_{\text{dir}} \). As shown in Column 3 of~\cref{fig:abalation}, the results demonstrated that while the T2I model struggled with highly localized edits, validating the challenge of using pre-trained T2I models for precise facial modifications. This emphasizes the necessity of \( \mathcal{L}_{\text{dir}} \) for enabling effective, localized edits in our framework.

Despite the limitations in performing precise edits, the model successfully preserved identity, attributed to the guidance provided by $\epsilon_1$.

Next, we removed the identity guidance from $\epsilon_1$, which was fine-tuned with \( \mathcal{L}_{\text{id}} \), and tested the model using only $\epsilon_2$, fine-tuned with \( \mathcal{L}_{\text{dir}} \). The results, as seen in Column 3 of~\cref{fig:abalation}, showed a substantial failure to preserve the original identity, which was corroborated by identity preservation metrics such as CosFace, VGGFace, and FaceNet. These metrics recorded significantly suboptimal values in this configuration (\cref{tab:identity}).

This ablation study highlights the critical importance of fine-tuning both $\epsilon_1$ and $\epsilon_2$ with \( \mathcal{L}_{\text{id}} \) and \( \mathcal{L}_{\text{dir}} \), respectively, to achieve accurate, identity-preserving localized edits.

\subsection{Mean Opinion Score Study}
We conducted a mean opinion score (MOS) study with 50 participants to qualitatively evaluate the realism and believability of our facial video edits, using a 1 to 10 rating scale (\cref{tab:mos}). A double-blind setup, with anonymized methods and randomized video order, ensured unbiased and robust assessment of our approach's perceived fidelity.

\subsection{Multiple Prompt-Localized Editing.}
Existing methods are constrained to performing one edit at a time on facial videos, necessitating successive processing for multiple edits and significantly increasing inference time. In contrast, our approach leverages the robust latent space of pre-trained SDE models. By embedding text-guided editing directly into the cross-attention layers of the fine-tuned U-Net~\cite{unet} $\epsilon_2$, we enable simultaneous application of multiple edits without added computational overhead. For instance, our method can process complex prompts such as ``[Add bangs, sunglasses, and a French beard to the face]" in a single step~\cref{fig:multiple_edits}.

\subsection{Generalization: Editing Faces in the Wild.}
Prior facial video editing works, trained end-to-end on curated datasets, tends to struggle when applied to in-the-wild facial videos(~\cref{fig:start} and Fig. \textcolor{red}{4} the results produced by DVA~\cite{Kim2022DiffusionVA} and STIT~\cite{tzaban2022stitch} ).
Our approach, however, harnesses the robustness of pre-trained SDE models trained on extensive data (e.g., LAION 2B dataset~\cite{schuhmann2022laion5b}), enabling effective editing of wild facial videos while preserving temporal consistency, even amidst challenging head poses and varied viewing angles as shown in ~\cref{fig:start} and Fig. \textcolor{red}{4} .

\section{Conclusion}
\label{sec:conclusion}
To address challenges such as high editing time, identity preservation, and achieving diverse edits in facial video editing, we proposed a novel pipeline that leverages pre-trained text-to-image (T2I) diffusion models, fine-tuned to maintain global identity while enabling directional edits. 

Furthermore, although stable diffusion is effective, it may introduce slight shifts in image texture, lighting,and  background. These shifts were observed in some of our results and should be considered in future work. Also further exploration for enhanced flexibility in editing facial videos through various guidance attributes, as well as a focus on both micro and macro expressions, presents promising advancements for creative applications.
\vspace{-2 mm}

\section{Acknowledgement}
\label{sec:Acknowledgement}
We would like to acknowledge partial support from IITM Pravartak Technologies Foundation. KM would like to acknowledge support from Qualcomm Faculty Award 2024.


 




{\small
\bibliographystyle{ieee_fullname}
\bibliography{egbib}
}

\end{document}